\documentclass[10pt,conference,anonymous]{IEEEtran}
\IEEEoverridecommandlockouts
\usepackage{cite}
\usepackage{amsmath,amssymb,amsfonts}
\usepackage{algorithmic}
\usepackage{graphicx}
\graphicspath{ {./figures/} }
\usepackage{textcomp}
\usepackage[table]{xcolor}  
\usepackage{color}
\usepackage{tcolorbox}
\usepackage{url}
\usepackage{colortbl}
\usepackage{caption}
\usepackage{multirow}
\usepackage{latexsym}
\usepackage{subfigure}
\usepackage{commath}
\usepackage{hhline}
\usepackage{epsfig}
\usepackage{graphics}

\def\BibTeX{{\rm B\kern-.05em{\sc i\kern-.025em b}\kern-.08em
    T\kern-.1667em\lower.7ex\hbox{E}\kern-.125emX}}
    
\newcommand{\chen}[1]{\textcolor{red}{Chen: #1}}

\begin{document}

\title{Robustness of on-device Models: Adversarial Attack to Deep Learning Models on Android Apps
}

\author{
\IEEEauthorblockN{Yujin Huang, Han Hu, Chunyang Chen\IEEEauthorrefmark{1}\thanks{*corresponding author}}
\IEEEauthorblockA{Faculty   of   Information   Technology, Monash University \\
	Melbourne, Australia \\
	yhua0096@student.monash.edu, han.hu@monash.edu, chunyang.chen@monash.edu}
}

\maketitle

\begin{abstract}
Deep learning has shown its power in many applications, including object detection in images, natural-language understanding, and speech recognition. 
To make it more accessible to end users, many deep learning models are now embedded in mobile apps.
Compared to offloading deep learning from smartphones to the cloud, performing machine learning on-device can help improve latency, connectivity, and power consumption. 
However, most deep learning models within Android apps can easily be obtained via mature reverse engineering, while the models' exposure may invite adversarial attacks.
In this study, we propose a simple but effective approach to hacking deep learning models using adversarial attacks by identifying highly similar pre-trained models from TensorFlow Hub.
All 10 real-world Android apps in the experiment are successfully attacked by our approach.
Apart from the feasibility of the model attack, we also carry out an empirical study that investigates the characteristics of deep learning models used by hundreds of Android apps on Google Play.
The results show that many of them are similar to each other and widely use fine-tuning techniques to pre-trained models on the Internet. 

\end{abstract}


\section{Introduction}
% \chen{When writing the latex, please follow the writing guide \url{https://github.com/ccywch/-LatexGuide}.}

Deep learning has shown its power in many applications, including object detection in images~\cite{pathak2018application}, natural-language understanding~\cite{young2018recent}, and speech recognition~\cite{nassif2019speech}.  
With the popularity of artificial intelligence, many mobile apps begin to incorporate deep learning inside for supporting advanced functionalities such as recommendation, face recognition, movement tracking, and translation. 
Many app development teams deploy their deep learning models onto the server, rendering the prediction to the app via the internet after taking the input from end devices. 
However, such a mechanism posts severe latency, extensive server resource requirement, heavy bandwidth load (e.g., streaming video), and privacy concern~\cite{dai2019machine}. 

Therefore, many apps begin to adopt on-device deep learning models, especially considering the increasing computing capability of mobile devices. 
Deploying deep learning models brings several advantages as follows.
First, there is no round-trip to a server, leading to the bandwidth saving, inference speeding up, and privacy preserving as privacy-sensitive data stays on the device.
Second, an Internet connection is not required, and apps can run in any situation.

Due to the benefits of embedding deep learning models into mobile apps, many popular frameworks begin to optimise their on-device frameworks, such as TensorFlow Lite (TFLite) of TensorFlow. 
Although many research works on improving model performance, few of them are concerned with model security, especially in terms of mobile apps. 
Unlike the central guardians of the cloud server, on-device models may be more vulnerable inside users' phones. 
For example, most model files can be obtained by decompiling Android apps without any obfuscation or encryption~\cite{xu2019first,chen2019gui}. 
Such model files may be exposed to malicious attacks.
Considering the fact that many mobile apps with deep learning models are used for important tasks such as financial, social or even life-critical tasks like medical monitoring or driving assistant, attacking the models inside those apps will be a disaster for users.

Over the last few years, many kinds of adversarial attacks~\cite{goodfellow2014explaining,kurakin2016adversarial,dong2018boosting,wang2018great,ma2018deepgauge} that attack deep learning models have been proposed.
Most of them are white-box attacks, i.e., knowing the structure and parameters of the deep learning models prior to the attack.
Although it is difficult to apply to a server-based model, which is almost a black box, the on-device model provides adversarial attacks with a chance to attack.
In this work, we design a simple but effective way to adapt existing adversarial attacks to hack the deep learning models in real-world mobile apps.

Given an app with a deep-learning model, we first extract the model and check if it is a fine-tuned model based on pre-trained models released by Tensorflow Hub~\cite{tensorflow}.
By locating its pre-trained model, we can train the adversarial attack approach based on that pre-trained model.
We apply the adversarial attack to the normal input (e.g., a picture waiting to be classified) and obtain the corresponding generated adversarial samples.
The experiment on 10 Android apps shows that all models are successfully attacked, with 23\% adversarial samples are wrongly classified. 

Apart from a pipeline for attacking the deep learning models in mobile apps, we also carry out an empirical study of the use of deep learning models within thousands of real-world Android apps.
We present those results by answering three research questions:
\begin{itemize}
	\item How similar are TFLite models used in mobile apps? (Section~\ref{section:Q1})
	\item How widely pre-trained TFLite models are adopted? (Section~\ref{section:Q2})
	\item How robust are fine-tuned TFLite models against adversarial attacks? (Section~\ref{section:Q3})
\end{itemize}

To this end, we collect 62,822 Android apps from Google Play. 
We identify 267 TFLite models among all the collected apps, which are developed by Google with the aim of optimising TensorFlow on edge devices.
Of these models, 87.64\% are structurally similar to each other, and 79.78\% have the same parameters as at least one other model.
Excluding 182 identical models, 61.18\% are more than 80\% similar to pre-trained models from TensorFlow Hub~\cite{tensorflow} in terms of their structural similarity, and 50.59\% have more than 43.93\% of the same parameters indicating that feature extraction and fine-tuning techniques are widely used on these pre-trained models.

Finally, we build a model attacker that can perform adversarial attacks on 10 representative fine-tuned TFLite models used in Android apps for experiments. 
Experimental results show that attacks with knowing fine-tuned models' pre-trained models are more than 3 times better than blind attacks in terms of the attack success rate. 
The results also show that the more similar the target model is to the pre-trained model, the more vulnerable it is to our targeted adversarial attacks.
We also discuss the Intellectual Property and defence of on-app models in Section~\ref{section:discussion}.

In summary, our contributions are as follows.
\begin{itemize}
    %\item We design and implement an analytical attack pipeline, including ModelDigger, ModelComparer, and ModelAttack, for analyzing TFLite DL models on Android apps. Capable of extracting TFLite DL models from Android apps, analyzing the relation between TFLite DL models, and testing the robustness of TFLite DL models against adversarial attacks. Our pipeline enables semiautomatic analysis of the security of on-device TFLite DL models.
    \item This is the first work to explore the feasibility of adversarial attacks in practical deep learning models from real-world mobile apps.
    
    %\item We apply our analytical attack pipeline on 62,822 Android apps collected from Google Play and Chinese app markets. Through the empirical analysis, we found that 79.78\% of on-device TFLite DL models have a relation with at least one model, and some are the same. The results indicate that intellectual property infringement is a practical problem for DL apps.
    \item We design and implement a complete pipeline, including ModelDigger, ModelComparer, and ModelAttack, to implement the attack on TFLite models in Android apps. We release our source code and detailed results\footnote{\url{https://github.com/Jinxhy/AppAIsecurity}} to the public and discuss both potential defences to those attacks and model IP protection.
    
    \item We also carry out an empirical study to show the latest use of deep learning models in the mobile industry, considering both model similarity and widely used fine-tuning approaches to developing deep learning models.
    %provide an analysis of the security of on-device TFLite DL models that adopt fine-tuned from the perspective of adversarial attacks. Our results show that fine-tuned TFLite DL models are unsafe, which implies that the security of on-device DL models needs to improve and should arouse the concern of mobile developers and researchers.
    
\end{itemize}

% \chen{We can remove this paragraph if we need more space.}
% The remainder of the paper is organized as follows. We describe
% the background knowledge in Section \ref{section:background}. We
% present our research goal and the analytical attack pipeline which helps us discover model relation, locate fine-tuned DL models and test the robustness of DL model in Section \ref{Workflow Overview}. The analysis results of model relation, fine-tuned DL models and the robustness of DL models are presented in Section \ref{section:Q1}, Section \ref{section:Q2} and Section \ref{section:Q3}, respectively. Section \ref{discussion} discuss the possible future work. Section \ref{related work} surveys the related work and Section \ref{conclusion} concludes the study.

% 1. why dl model in mobile apps
% 2. issues with dl model in mobile apps
% 3. related works about adversarial attacks and app security
% 4. statement of our approach

% The contributions of this work:
% \begin{itemize}
% 	\item an empirical study of Dl models on Android apps.
% 	\item an approach to attack DL models practically on Android apps.
% 	\item evaluation for demonstrating the approach efficiency.
% \end{itemize}

% \chen{One contribution is that we release a set of open-source tools for analyzing the deep learning models on Android Apps such as ModelDiggger, ModelComparaer, ModelAttacker}
\section{Background}
\label{section:background}

\subsection{On-device Deep Learning Model}
Being ubiquitous, mobile devices are among the most promising
platforms for deep learning. In practice, deep learning model inference can be offloaded to the cloud or executed on mobile devices. 
The cloud-based deep learning model requires mobile devices to send data to the server then retrieve inference results, which comes with multiple drawbacks, such as low privacy protection, high latency, and extra costs for cloud providers. 
In comparison, the on-device deep learning model executes inference on smartphones without uploading data, which improves user privacy, eliminates the impact of network latency, and lowers cloud costs for mobile developers.

On-devcie deep learning models are typically implemented through deep learning frameworks such as Google TensorFlow\cite{tensorflown} and TFLite\cite{tensorflowlite}, Facebook PyTorch\cite{pytorch} and Caffe2\cite{caffe2}, Tencent NCNN\cite{ncnn}, and Apple Core ML\cite{coreml}. 
The availability of these practical frameworks allows mobile app developers to produce on-device deep learning models without tremendous engineering efforts. 
TensorFlow and TFLite frameworks, which run TF models on mobile, embedded, and IoT devices contribute almost 40\% of the total number of deep-learning-based mobile apps in 2018\cite{xu2019first}. 
TFLite is the most popular technology used to run deep learning models on smartphones, as it has GPU support and has been extensively optimised for mobile devices\cite{sanabria2018code}. 
However, pre-trained deep learning models are commonly utilized on mobile systems due to the high overheads incurred when training a new deep learning model from scratch\cite{ccurukouglu2018deep}. Through the use of pre-existing deep learning models, mobile app developers can directly perform inference with low latency and low power consumption on mobile devices.

\subsection{Adversarial Attack}
Deep neural networks are highly vulnerable to adversarial attacks, which add subtle perturbations to inputs that lead a deep learning model to predict incorrect outputs with high confidence level\cite{zhang2019adversarial}. 
Suppose that there is a deep learning model $M$ and an original example $x$ that can be correctly classified, i.e., $M(x)=y_{true}$, where $y_{true}$ is the true label of $x$. By adding a subtle perturbation to $x$, attackers can produce an adversarial example $x^{'}_{adv}$ that is highly similar to $x$ but which is misclassified by model $M$, i.e., $M(x^{'}_{adv})\neq y_{true}$\cite{szegedy2013intriguing}. 
The whole process can be summarised as:
\begin{equation}
    x^{'}_{adv} = x + \epsilon * attack_i(\nabla J (\theta,\ x,\ y_{true}))
\end{equation}
where $y_{true}$ represents the original label or class of input $x$, $\epsilon$ represents a multiplier to ensure the perturbations are small, $attack_i$ represents the $i_{th}$ type of adversarial attack, $\theta$ represents parameters, and $J$ is the loss.

\begin{figure}[htbp]
\centering
\subfigure[Original Input Image] {
    \begin{minipage}{3cm}
    \centering
    \label{fig:origin input}
    \includegraphics[width=3cm]{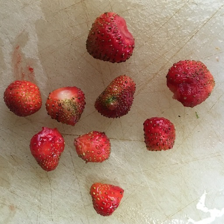}
    \end{minipage}
}
\subfigure[Adversarial Example] {
    \begin{minipage}{3cm}
    \label{fig:adversarial examples}
    \centering
    \includegraphics[width=3cm]{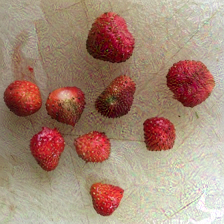}
    \end{minipage}
}
\caption{Comparison of original image and adversarial example}
\label{fig}
\end{figure}

The differences between adversarial examples and standard inputs are always so subtle that humans cannot even notice the modification, yet the model can still be fooled and misclassified. 
For example, Figure~\ref{fig:origin input} is an input image, and Figure~\ref{fig:adversarial examples} shows an adversarial example of Figure~\ref{fig:origin input}.

%In most case, the the $L_p$ norm of the perturbation (i.e. adversarial noise) is required to construct adversarial examples and should be less than an specified value $\epsilon$ as $\|{x^{'}-x}\|_{p}\leq\epsilon$, where $p$ could be $0, 1, 2, \infty$\cite{dong2018boosting}.

%Undoubtedly, it is a hazardous signal that adversarial examples can successfully fool the model. Once the attack is successful, the attacker can use adversarial examples to weaken or even change the classifiers' ability without the access to the underlying features, and then use the leaked model to carry out further illegal measures, causing unassessable losses.

%In this paper, we analyzed the common machine learning models in the Android markets, and selected eighteen representative types of adversarial attacks to check the robust of fune-tuning TFLite DL models. 

Adversarial attacks can mainly be divided into the Black-box Attack and the White-box Attack, or the Targeted attack and the Non-target attack. 
Several  representative adversarial attack approaches that fall into these categories are: Fast Gradient Sign Method (FGSM)~\cite{goodfellow2014explaining}, Projected Gradient Descent (PGD)~\cite{aleks2017deep}, Basic Iterative Attack (BIM)~\cite{kurakin2016adversarial}, Decoupling Direction and Norm Attack (DDN)~\cite{Rony_2019}, Deep Fool Attack~\cite{Moosavi_Dezfooli_2016}, Newton Fool Attack~\cite{newtonFoolAttack}, Inversion Attack~\cite{inversion_attack}, and Boundary Attack~\cite{brown2017adversarial}. 
These approaches will be used to access the robustness of deep neural networks.
% \chen{Please tell more about these attacks including the basic principles}
FGSM uses the gradients of the neural network to create adversarial examples,
while PGD is an iterative extension of FGSM that performs a projected gradient descent.
BIM is also is a straightforward extension of FGSM, iteratively employing an attack multiple times.
DDN performs attacks by decoupling the direction and the norm of the adversarial perturbation added to the image.
Deep Fool Attack works by projecting the input onto the nearest non-linear decision boundary,
while Newton Fool Attack works by performing a gradient descent that declines the probability of the original class.
Inversion Attack involves the attacker extracting information related to training data, such as some sensitive training-data characteristics, from model prediction results.
Boundary attack only focuses on output class queries, and starts attacking with a sizeable adversarial perturbation, then seeks to reduce the perturbation while staying adversarial.

\section{Workflow Overview}
\label{section:workflow overview}

\begin{figure*}[htpb]
\centering
\includegraphics[scale=0.5]{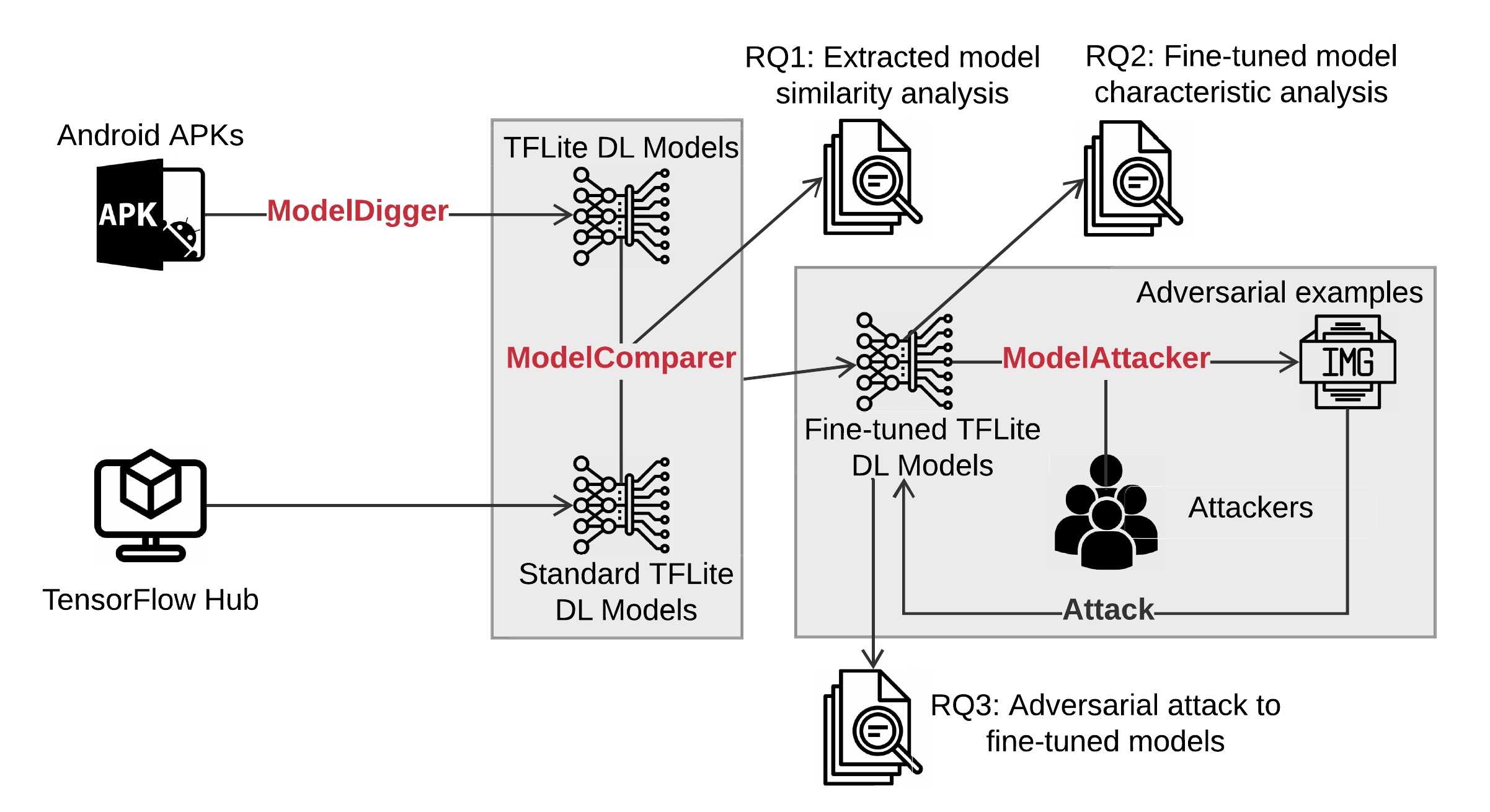}
\caption{The overall workflow of the analytical attack pipeline.}
\label{fig:structure}
\end{figure*}

We design and implement an analytical attack pipeline to enable our research goals on plenty of Android apps. This pipeline can (i) discover the relation between TFLite models; (ii) locate fine-tuned TFLite models; (iii) evaluate the robustness of TFLite models against adversarial attacks. Our pipeline runs in a semiautomatic way (the overall workflow, is illustrated in Figure~\ref{fig:structure}).

For the preparation of our study, we crawled 62,822 mobile apps from Google Play across various categories (e.g., Photography, Social, Shopping) related to the image domain.

The very first step in our pipeline is to identify TFLite deep learning apps and extract TFLite models among a given set of Android APKs as input. 
This is achieved via the tool named ModelDigger, which first disassembles APKs to the nearly original form then obtains the source codes and asset files from the APK archive via Apktool\cite{apktool}. 
After decompiling the collected APKs, the ModelDigger examines whether a decomposed APK comprises files with the TFLite model naming convention\cite{sun2020mind}, if any such exist, it labels this APK with a tag then extracts the TFLite models from it. 
During extraction, ModelDigger checks the completeness and operability of each model by loading a model and performing inference on randomly generated data, then filters out unavailable models to ensure the quality of the extracted TFLite models.

The analytical attack pipeline then discovers the relation between the extracted TFLite models and locates the fine-tuned models among them, which is achieved by the tool called ModelComparer. The core purpose of ModelComparer is to detect the longest common sub-layers between two TFLite models, then calculate the structural and parametric similarities between them. 

% After computing the similarities for each extracted TFLite DL model, it performs analysis to discover the relation between each extracted TFLite DL model and determine which of them are fine-tuned by taking advantage of standard pre-trained TFLite DL models.

Finally, to evaluate the adversarial robustness of fine-tuned TFLite models, we manually collect model-based input data for each fine-tuned TFLite model and enter them into the adversarial example generator, ModelAttacker, which performs various adversarial attack algorithms based on the architecture of a TFLite model. To ensure the accuracy of an evaluation, the adversarial examples are generated using a set of pre-set attack parameters. This guarantees that the generated adversarial examples are humanly imperceptible.

\textbf{Initial results} We identify 101 TFLite deep learning apps and successfully extract 267 TFLite models. The results are shown in Table~\ref{table: 1}. As observed, 40.45\% TFLite models (108 out of 267) come from the Photography category, while 9.00\% of them (24 out of 267) belong to the Business category. It is evident that most TFLite deep learning apps contain more than one model. The reasons for this could be (i) the TFLite deep learning apps have two or more deep-learning-based functionalities; or (ii) the tasks that the TFLite deep learning apps perform require multiple models to work together. In addition, we find that 95.51\% extracted TFLite models (255 out of 267) are convolutional neural network (CNN) models, and the remaining 12 models are recurrent neural network (RNN) models. The CNN models are proficient at capturing visual characteristics from images and are commonly used for image processing and classification. Hence, the results are consistent with our intention of app collection: most TFLite models are commonly used in task domains that are related to the image.

We will discuss ModelComparer and ModelAttacker in further detail in Sections \ref{section:Q1} and \ref{section:Q3}, respectively.

% As part of our analytical attack pipeline, ModelDigger takes Android APKs as input and outputs that use the TensorFlow Lite framework and their TFLite DL models. ModelDigger first disassembles Android APK files and extracts the app asset files and the native libraries via Apktool\cite{apktool}. To find TFLite model files, previous work\cite{xu2019first} only scans the assets folder of a decomposed DL app with the conventional model naming schemes like ModelName.tflite, however, we find that many TFLite DL model files are not only stored outside the assets folder but also do not follow the clear naming convention. Hence, inspired by Sun et al.\cite{sun2020mind}, we use a novel search approach combining three suffixes, including ".tfl", ".lite", and ".tflite" to search each folder of a decomposed DL app thoroughly.

% After locating the TFLite DL models, ModelDigger performs extraction on them. However, we here face the challenge that some TFLite DL models are broken or unavailable. Therefore, to ensure the quality of the extracted TFLite DL models, ModelDigger first filters out the model files whose size is more significant than 16 KB then
% loads each filtered TFLite DL model via TensorFlow Lite framework\cite{tensorflowlite}. If a TFLite DL model is loaded without any exception, we argue that this model is complete and operational. Consequently, ModelDigger extracts all validated TFLite DL models and passes them to ModelComparer to explore their relations.

\begin{table}
\centering

\caption{Numbers of TFLite DL apps and models.}
\label{table: 1}

\setlength{\tabcolsep}{20pt}
\renewcommand{\arraystretch}{1.1}
\resizebox{\linewidth}{!}{%
\begin{tabular}{l|c|c} 
\noalign{\hrule height 1pt}
\textbf{Category} & \textbf{\#TFLite App} & \textbf{\#DL Model} \\ 
\noalign{\hrule height 1pt}
Photography & 34 & 108 \\ 

Productivity & 23 & 37 \\ 

Social & 16 & 41 \\ 

\textcolor[rgb]{0.2,0.2,0.2}{Automation} & 10 & 30 \\ 

\textcolor[rgb]{0.2,0.2,0.2}{Shopping} & 9 & 27 \\ 

\textcolor[rgb]{0.2,0.2,0.2}{Business} & 9 & 24 \\ 
\noalign{\hrule height 1pt}
\textcolor[rgb]{0.2,0.2,0.2}{Total} & 101 & 267 \\
\noalign{\hrule height 1pt}
\end{tabular}}
\end{table}
\section{Q1: how similar are TFLite models used in mobile apps?}
\label{section:Q1}

%As mentioned in Section \ref{section:background}, pre-trained DL models are commonly used in mobile devices, as training a new DL model from scratch is time-consuming. For ease of development of on-device DL models, TensorFlow official provides a comprehensive repository that contains a series of pre-trained DL models to developers for fine-tuning or reuse\cite{tensorflow}. Hence, there may be some intrinsic connections between on-device DL models. In this research question, we investigate into how on-device TFLite DL models are similar in DL apps (i.e. model relation).
Since many apps contain deep learning models with similar functionalities, as mentioned in Section \ref{section:workflow overview}, we are investigating whether they use the same or similar deep learning models within their apps.
% \chen{Do not talk about pre-training in this section, and that is what we are exploring in the next RQ.}

\subsection{ModelComparer: extracting the similarity between models}
\begin{figure*}
\centering
\includegraphics[width=\linewidth]{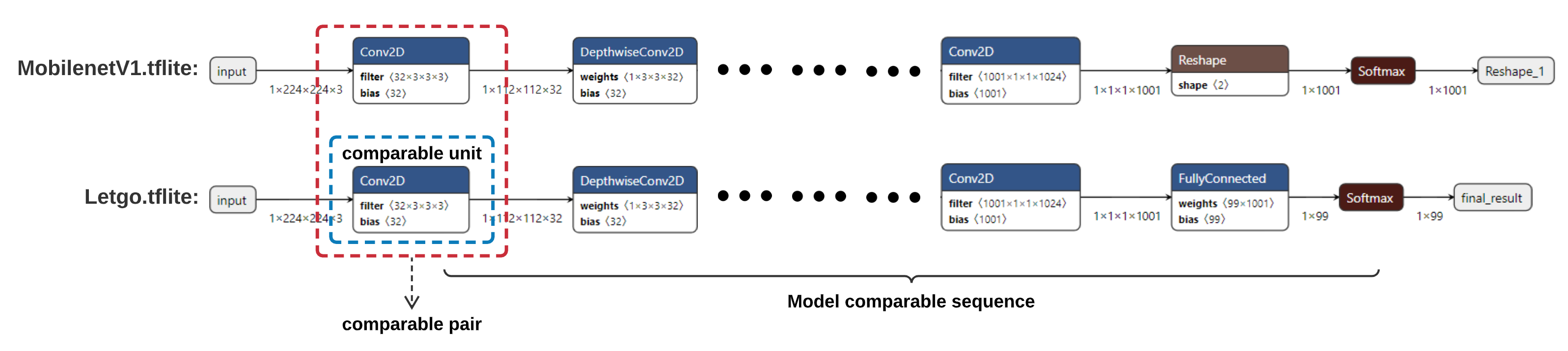}
\caption{TFLite DL model conversion in ModelComparer, visualized by Netron\cite{netron}.}
\label{fig:conversion}
\end{figure*}

%As one of our analytical attack pipeline's core parts, ModelComparer first takes on-device TFLite DL models that we have already extracted as input and outputs the similarity between the extracted models. Subsequently, ModelComparer performs a similar comparison for the extracted TFLite DL models and standard pre-trained TFLite DL models from TensorFlow Hub and outputs which of extracted models are fine-tuning. To discover the relations between on-device TFLite DL models, ModelComparer first computes the structural similarity between two TFLite DL models. Through our experiments and observations, we found that if the structural similarity between the two models is equal to or greater than 80\%, they are similar in terms of structure. After computing the structural similarity, ModelComparer would further calculate the parameter similarity for those models with structural similarity equal to or greater than 80\% to verify whether they have a relation. For identifying the fine-tuning TFLite DL models, the comparison procedure is slightly different, which would be discussed in Section \ref{section:Q2}.
When evaluating the similarity between models, we calculate two metrics: the structural similarity and the parametric similarity.
Each deep learning model consists of multiple layers, like one convolutional model containing several convolution layers and pooling layers.
Given one deep learning model extracted from the mobile app, to facilitate comparison we first convert it to a sequence of elements, with each element representing one layer of the model. 
One unit in the sequence contains one layer's information, including identifier, shape, and data type.
As seen in Figure~\ref{fig:conversion}, one convolutional layer is encoded to MobilenetV1/Conv2d\_0/Relu6,[1,112,112,32],uint8.
After converting models to sequences, we detect the longest common subsequence between any two models $M1, M2$.
We then further calculate the structural similarity of them by:
%To compute the structural similarity, ModelComparer converts extracted TFLite DL models into comparable sequences in form of text. As an example, the structure comparison of two TFLite DL models, MobilenetV1 and Letgo, as illustrated in Figure~\ref{fig:conversion}. A model comparable sequence consists of multiple comparable units, each unit represent one layers in the model. A comparable unit contains layer's information, including identifier, shape, and data type. For instance, the layer inside the blue box is converted to the text form: MobilenetV1/Conv2d\_0/Relu6,[1,112,112,32],uint8. ModelComparer treats two corresponding comparable units, i.e., the layers in the same position in two models, as a comparable pair. Based on the comparable pairs for each extracted TFLite DL model, ModelComparer can compute each extracted TFLite DL model's structural similarity by using the text-matching algorithm\cite{bergroth2000survey} to calculate the proportion of the longest common sub-layers between the model itself and the rest of the extracted models. The formula for similarity measurement is:
\begin{equation}
    similarity(M1, \ M2) = \frac{2 * L_{match}}{L_{total}}
\end{equation}
where $L_{match}$ is the number of longest common subsequence in two models, and $L_{total}$ is the total number of two models' layers. 
%$target$ is the target model for measurement, and $remanent$ is one of all extracted models except the target one. 
Note that one unit in the sequence is counted as matched to one unit in the other sequence only if the attributes of them are totally the same.
The similarity score is in the range of 0 to 1, and the higher the similarity score, the more structurally similar the two models are.

Apart from the structural similarity, we further adopt the parametric similarity to determine the model similarity.
We still convert the model into a sequence, with each layer as one unit, but we adopt the detailed parameter of that layer as the attribute. 
Given two models, $M1, M2$, with a high structural similarity (greater than or equal to 0.8), we perform a subtraction on each corresponding unit, i.e., two units in the same position, as two models have a similar structure. 
If the calculation result is zero, the Boolean value True that represents two units' parameters are the same will be stored in a sequence with the order. If not, False will be stored. 
We then compute the parametric similarity of them by:
\begin{equation}
    similarity(M1, \ M2) = \frac{N_{True}}{N_{total}}
\end{equation}
where $N_{True}$ is the number of longest continuous True values, and $N_{total}$ is the total number of Boolean values in the result sequence. 
The similarity score is in the range of 0 to 1, and the higher the similarity score, the more parameters are similar between the two models.

%To compute the parameter (weight and bias) similarity for those TFLite DL models with structural similarity equal to or higher than 90\%, ModelComparer performs a similar process to the structure comparison. In the parameter comparison, ModelComparer also converts TFLite DL models to comparable sequences, but the comparable units store a layer's parameters instead of its structure information. Subsequently, ModelComparer performs subtraction for each comparable pairs in every two models (i.e. $target$ and $remanent$) to check the parameter equality and records the results based on the order of comparable pairs. Finally, ModelComparer computes the parameter similarity between every two models by calculating the percentage of the longest common sub-layers with the same parameters between the target model and the rest of the extracted models. 
% \chen{Do not understand how do you calculate the similarity, please refer to my Section 3.3. in my previous paper \url{https://chunyang-chen.github.io/publication/proactiveEdit_cscw18.pdf}}

To further analyze the overall relationships among models from different apps, we model their similarity in a single graph.
We take each model as the node and their relationship as the edge in the graph.
Note that according to our observations, there is one edge between models if both the structural and parametric similarities are higher than 0.8.
We then carry out community detection to group highly related nodes in the graph. In graph theory, a set of highly correlated nodes is referred to as a community (cluster) in the network. In this work, we use the Louvain method~\cite{blondel2008fast} implemented in the Gephi~\cite{bastian2009gephi} tool to detect communities.
We visualise different communities in different colors and the edges with higher similarity as large thicknesses, as seen in Figure~\ref{fig:relation}.
For each model, we annotate its app name along with its extracted order over the node in the graph visualisation.

\subsection{Results}
%\textbf{$\bullet$ Are on-device TFLite DL models similar?} 
Our study shows that 87.64\% of on-device TFLite models (234 out of 267) are structurally highly similar, i.e., the similarity score is equal to or higher than 0.8. 
Among the 234 structurally similar TFLite models, 91.03\% (213 out of 234) have the same parameters (i.e., 100\% parametric similarity) to at least one model.
Specifically, 182 TFLite models are identical, both in terms of their structure and parameters.
These models are directly using the Google Mobile Vision framework\cite{googlemobilevision}. 
%Thus, we deem the reuse of pre-trained models generated by deep learning frameworks is a possible reason for identical models.
Apart from these identical models, we also find that 52 TFLite models across various app categories, such as photography beauty, business, and productivity, are not only structurally similar to models belonging to the same category but also have relations to the models in other categories. 
The results indicate that most on-device TFLite models have relations, especially those TFLite models that fall within the same app category.
%  \chen{please tell that how many models are totally the same 100\% and analyze the reasons.}
%  \chen{??There are so many highly similar models, why are there only tens of them in the Fig~\ref{fig:relation}?}.

% \chen{Just 3 paragraph including clusters, models similarity within clusters and model connection across clusters}

%\textbf{$\bullet$ How are on-device TFLite DL models similar within and outside their task domains (clusters) ?} To investigate the relations between on-device TFLite DL models in different task domains, we manually classify the domain of each extracted TFLite DL model (exclude 182 identical models developed by Google) according to its corresponding app description and app content. The results are shown in Table~\ref{table: 2}. Overall, image classification, object detection, and image segmentation are the most popular application area of on-device TFLite DL models, far more than other domains (Text recognition, Face detection, Speech recognition, etc). 
% This reflects that the field of computer vision is still a center in DL-based mobile apps and the security of DL models regrading this field should raise concerns.
  
\begin{figure*}
\centering
\includegraphics[scale=0.78]{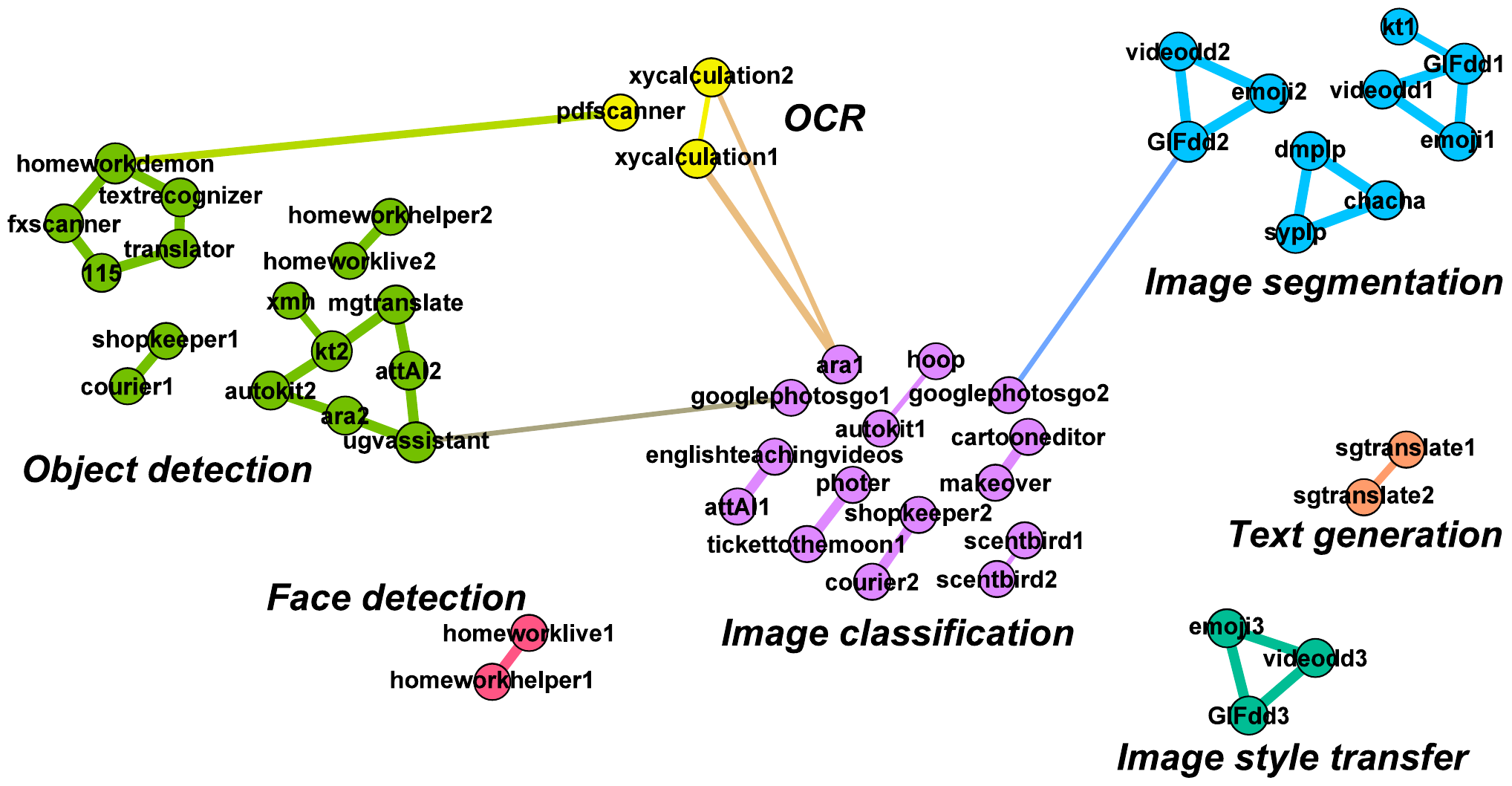}
\caption{Relations between TFLite models}
\label{fig:relation}
\end{figure*}

As seen in Figure~\ref{fig:relation}, there are 7 main clusters of on-device deep learning models such as image classification, segmentation, OCR, object detection, which are widely used in our daily lives.
Most deep learning models in computer vision are very similar to each other in terms of both structure and parameters, while textual models are mostly different.
According to our further analysis of those highly-similar models, we find that 75\% of them are using the same model structure as MobileNet~\cite{howard2017mobilenets,sandler2018mobilenetv2} which is a lightweight deep neural network specifically developed for the mobile platform by Google.
We also find that 7 model pairs are identical. The image classification and object detection clusters contain 4 and 2 pairs, while the face detection only contains one.
As observed, each model pair's models come from different apps as the node name represents which app a model belongs to.
This may be caused by using the same pre-trained model or by stealing models from similar apps.

%all TFLite DL models within the image style transfer cluster are entirely the same (i.e. 100\% similarity), which forms a novel sub-cluster. Similarly, there also are 2 and 3 sub-clusters in the object detection and image segmentation clusters, respectively. The possible reason for that is TFLite DL models belonging to the same task domain utilize the same standard pre-trained TFLite DL models without modification or directly reuse models of similar DL apps. Apart from that, the rest of TFLite DL models residing in the same clusters (image classification, face detection, OCR, etc) are paired in pairs. For the 11 model pairs, 6 of them come from the image classification cluster, while object detection, OCR, face detection, and text generation clusters only contain either 2 or 1. Specifically, 8 model pairs have 100\% similarity, i.e., two models are the same in terms of structure and parameter, and the remaining 3 have 80\% similarity. This implies that the model relations within the image classification clusters are stronger than other clusters.

Some clusters share connections, such as object detection with image classification and OCR with image classification.
Note that many clusters share a connection with image classification models, as image classification is the backbone of these advanced tasks.
For example, one model called googlephotosgo2 from the app google photogo adds one more layer to the model GIFdd2 from the app GIF dd, while other layers are totally the same.
Nevertheless, for other advanced tasks like image style transferring or text generation, they are rarely related to other clusters due to their specificity.

\begin{center}
\begin{tcolorbox}[colback=black!5!white,colframe=black!75!black]
\textbf{Summary:}
On-device deep learning models are widely used in mobile apps covering a set of common tasks like image classification, object detection etc.
Many models are similar in terms of both structure and parameter value, especially those for image classification tasks.
Some of them directly copy the models from other apps or add just several layers beyond others' models.
%On-device TFLite DL models with high similarity typically appear in the same task domains, which is likely due to the adoption of the same pre-trained DL model or the reuse of the similar DL model by others. On the other hand, on-device TFLite DL models in different task domains also have similarities, and most of the cross-cluster model relations are related to the image classification clusters.
\end{tcolorbox}
\end{center}
%\section{Q2: how widely fine-tuned TFLite models are adopted?}
\section{Q2: How widely pre-trained TFLite models are adopted?}
\label{section:Q2}
Many deep learning models across different mobile apps are very similar, as mentioned in last section, and many layers between them are almost the same except the last few layers.
That may be because of the widely used fine-tuning techniques~\cite{pan2009survey} in deep learning area i.e., adding or adjusting final layers in the pre-trained models.
That fine-tuning approach is especially useful for application with few labelled data which make it popular. 
In this section, we further verify if models in mobile apps are fine-tuned models based on pre-trained models.
% \begin{comment}
% For ease of development of deep learning models for mobile devices, TensorFlow official provides a comprehensive repository that contains a series of pre-trained models to mobile developers, which facilitates the implementation of deep learning in mobile apps\cite{tensorflow}. 
% To investigate whether a TFLite model takes advantage of a standard pre-trained model, we collect 91 state-of-the-art pre-trained models across different task domains from TensorFlow Hub for comparison.
% \end{comment}

\subsection{Extracting Similarity between App Models and Pre-trained Models}
A pre-trained model is a saved network that was previously trained on a large general dataset. 
Instead of building a model from scratch to solve a similar problem, developers then either use the pre-trained model as is or use transfer learning to customise this model to a given task.
There are some commonly used pre-trained models, such as  ResNet~\cite{he2016deep} on ImageNet~\cite{deng2009imagenet} for image classification and BERT~\cite{devlin2018bert} on Wikipedia~\cite{moviebook} for language modeling.
TensorFlow Hub (TensorHub)~\cite{tensorflow} is a repository of pre-trained machine learning models developed by Google.
There are 91 models including most common ones such as MobileNet \cite{howard2017mobilenets,sandler2018mobilenetv2}, Inception \cite{szegedy2016rethinking}, and SqueezeNet\cite{iandola2016squeezenet}.

Given the collected 85\footnote{Excluding the 182 identical models from the last section.} models from practical mobile apps, we check the similarity of each model in our collection with models in TensorHub in terms of the structural and parametric similarity, as mentioned in Section~\ref{section:Q1}.
Note that for each model in our collection, we only locate the most similar one from TensorHub.
Apart from the quantitative analysis, we also manually check the models with high similarity to the pre-trained models in TensorHub for investigating their difference.

\subsection{Results}

\begin{figure}
    \centering
    \includegraphics[width=\linewidth]{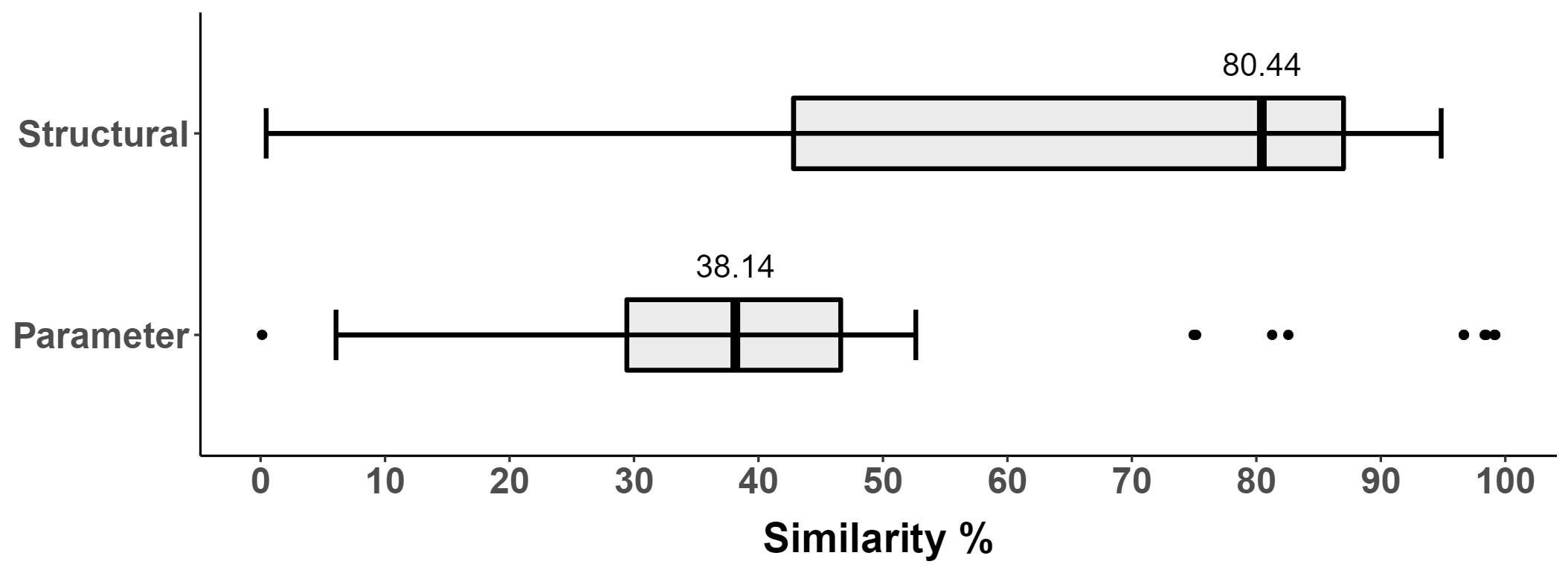}
    \caption{Distributions of similarities of fine-tuned and pre-trained models.}
    \label{fig:fine_tuning_approach}
\end{figure}

Figure~\ref{fig:fine_tuning_approach} shows the similarity distribution of models from mobile apps and those from TensorHub.
In terms of structural similarity, we can see that 43 (50.59\%) models from our collection are more than 80\% similar to those in TensorHub, as the pre-trained models have been optimised and cover common task domains like image classification.
It represents that many models in apps are sharing similar architecture.
We then manually check the pre-trained models which are highly similar to the models in mobile apps.
As seen in Table~\ref{table:pretrain}, pre-trained MobileNet V1 and V2 are the most popular ones (21/43) which is specifically customised for mobile devices in computer-vision related tasks.
Most similar pre-trained models are about image related tasks like image classification, object detection or segmentation.
In contrast, there are no highly similar pre-trained models about text or speech.

In terms of the parametric similarity, there are always some layers in the app model with the same parameters to that of pre-trained model in TneforHub as seen at Figure~\ref{fig:fine_tuning_approach} shows.  
Specifically, for 43 models with more than 80\% structural similarity, 6 (13.95\%) of them are of more than 95\% parameter value similar.
These models use the representations learned by a previous pre-trained network to extract meaningful features from new samples and this kind of fine-tuning can be called feature exatraction~\cite{pan2009survey}. For instance, there are 66 layers in the pre-trained MobileNet V1, freeze all layers of the base model and only train the parameters of the newly added classifier. This allows to repurpose the feature maps learned previously by the base model trained on a larger dataset in the same domain.

% \chen{Can you please give some detailed examples? for example, there are xxx layers in the pretrained MobileNet, Unfreeze a few of the top layers of a frozen model base and jointly train both the newly-added classifier layers and the last layers of the base model. This allows to "fine-tune" the higher-order feature representations in the base model in order to make them more relevant for the specific task.}

% \begin{comment}
% fine-tuned TFLite models adopt Fine-Tuning, while few of them use Feature Extraction.
% According to our further analysis of 37 (86.05\%) models that adopt Fine-Tuning, we find that 83.78\% (31 out of 37) of them retrain more than 50\% of their parameter as they are dedicated to specific tasks regarding their apps, i.e., more accurately classify the subclasses of one or more labels of the pre-trained model.
% Compared to Fine-Tuning, Feature Extraction (13.95\%, 6 out of 43) is less commonly used as the task domain of the fine-tuned model is limited by the pre-trained model, i.e., the learned parameters of the pre-trained model cannot be changed. 
% \end{comment}

\setlength{\tabcolsep}{1pt}
\renewcommand{\arraystretch}{1.2}
\begin{table}
\centering
\captionsetup[table]{format=plain,labelformat=simple,labelsep=period}
\caption{Numbers of fine-tuned TFLite DL models.}
\label{table:pretrain}
\resizebox{\linewidth}{!}{%
\begin{tabular}{l|c|c} 
\noalign{\hrule height 1pt} \multicolumn{1}{c|}{\textbf{Task domain}} & \textcolor[rgb]{0.2,0.2,0.2}{\textbf{Pre-trained model type}} & \textcolor[rgb]{0.2,0.2,0.2}{\textbf{Fine-tuning model count}} \\ 
\noalign{\hrule height 1pt}
\multirow{3}{*}{Image classification} & MobileNet V1 and V2 &  21 \\
%\cline{2-3}
 & InceptionV3 & 2 \\
%\cline{2-3}
 & SqueezeNet & 1 \\ 
\hline
\multirow{2}{*}{Object detection} & COCO SSD MobileNet v1 &  9 \\
%\cline{2-3}
 & Google Mobile Object Localizer & 2 \\ 
\hline
Image segmentation & DeepLabv3 & 7 \\ 
\hline
Pose estimation & PoseNet & 1 \\ 
\noalign{\hrule height 1pt}
Total &  & 43 \\
\noalign{\hrule height 1pt}
\end{tabular}}
\end{table}

\begin{center}
\begin{tcolorbox}[colback=black!5!white,colframe=black!75!black]
\textbf{Summary:}
Many mobile apps fine-tune the pre-trained deep learning models for their own purpose especially in computer-vision related tasks.
The fine-tuned models are similar to the pre-trained models in TensorHub in terms of both structure and parameters.
%Fine-tuned deep learning models gains popularity among deep learning mobile apps especially in the field of computer vision. Many models adopt Fine-Tuning approach for retraining.
%Convolutional neural network is dominant on the fine-tuned models.

\end{tcolorbox}
\end{center}

\section{Q3: how robust are fine-tuned TFLite models against adversarial attacks?}
\label{section:Q3}

\begin{table*}[ht]
  \centering
  \caption{Details of selected 10 models}
    \begin{tabular}{c|c|c|c|c}
    \noalign{\hrule height 1pt}
    \textbf{ID} & \textbf{App Name} &\textbf{Model Name} & \textbf{Similarity} & \textbf{Model Function} \\
    \noalign{\hrule height 1pt}
    1     & converted\_model.tflite & QQ browser & 74.96 & Identify plants \\
    2     & graph.tflite & Fresh Fruits & 29.79 & Determine if the fruit has gone bad \\
    3     & mobilenet.letgo.v1\_1.0\_224\_quant.v7.tflite & Taobao & 47.02 & Identify products \\
    4     & mobilenet\_v1\_1.0\_224\_quant.tflite  & Image Checker & 48.03 & Image classification of Imagenet database\\
    5     & optimized\_graph.tflite & iQIYI & 45.55 & Identify actors\\
    6     & pothole\_detector.tflite & Tencent Map & 44.97 & Judge traffic conditions \\
    7     & mobile\_ica\_8bit\_v2.tflite & Bei Ke & 81.27 & Identify scenes\\
    8     & model.tflite & Baidu Image & 46.61 & Identify flowers\\
    9     & pokemon\_mobilenetv2.tflite & My Pokemon & 96.67 & Identify different Pokemon\\
    10    & skin\_cancer\_best\_model.tflite & Palm Doctor & 98.46 & Identify different skin cancers\\
    \noalign{\hrule height 1pt}
    \end{tabular}%
  \label{tab:model_detail}%
\end{table*}%

The latter 2 research questions demonstrate that the pre-trained models and fine-tuning approach are widely used in developing real-world on-device deep learning models.
Considering the popularity of adversarial attacks on neural networks and the availability of identifying the pre-trained models, we further explore the possibility of applying off-the-shelf adversarial attack to existing deep learning models inside mobile apps.
% \chen{may tell more why adversarial attacks are white-box why it needs the pre-trained model?}
Since most adversarial attacks are white-box based, i.e., FGSM and BIM, relying on the understanding of target models including the training data and runnable for adjusting its parameter, we propose a method called ModelAttacker to attack the deep learning model based on the identification of its pre-trained model.  
Therefore, we present the performance of our targeted attack compared with blind attack in real-world mobile apps from Google Play in this section.
\begin{comment}
This question aims to explore how robust are fine-tuning TFLite DL models against adversarial attacks.
As we all know, DL models' ability is the core competitiveness and intellectual properties of DL apps. The consequences of failed and leaked DL models are profound and wide. 
In the fields of online shopping or medical treatment, attackers can access our privacy, induce us to shop, make unwise decisions, or even steal our money directly.
Moreover, the impacts are far more than unevaluable economic losses. If the security of DL models can not be guaranteed, then deep learning can never be used in some areas with strict security requirements, such as finance, which will significantly restrict the whole research field's development. 
Adversarial attack is one of the most common approaches to attack machine learning models.
To answer this question, we build a tool called ModelAttacker to check the robustness of fine-tuning DL models by employing multiple kinds of adversarial attacks.

In this section, the section begins with an introduction to the methodology of ModelAttacker. 
Then, the evaluation of ModelAttacker will be detailed.
Finally, findings and results will be discussed. 
\end{comment}

\subsection{Methodology of ModelAttacker}
% \chen{why are you making the procedures so complicated? Please check if my writing is correct or not.}
Given a deep learning model decompiled from the mobile app, we first compute its similarity with all models from TensorHub by the ModelComparer mentioned in Section~\ref{section:Q1}.
According to the similarity ranking, we select the most similar pre-trained model from our database in terms of structure and parameter value.
If the similarity is above the threshold 80\%, which is experimentally set up, we train the parameters of adversarial attacks on the pre-trained model.
We then apply the adversarial attack to generate adversarial examples to feed into the target deep learning models.

As mentioned in Section~\ref{section:background}, there are 11 representative attacks adopted from 4 all main categories to carry out experiments.

\subsection{Evaluation of ModelAttacker}
\subsubsection{Evaluation of attacking fine-tuned models} 
% \chen{1) Please also introduce the evaluation metric that you use; 2)how many adversarial attack pictures do you use for each model?}
As shown in Table~\ref{table:pretrain}, there are 43 TFLite models from Google Play found to be fine-tuned.
We pick up 10 representative models that are all fine-tuned from MobileNet V1 and MobileNet V2 as they are most commonly used.
Table~\ref{tab:model_detail} show the detail of selected 10 models.
The \emph{Similarity} indicates the similarity between the fine-tuned model and its pre-trained model.
We utilize ModelAttacker to employ selected 11 kinds of adversarial attacks on these TFLite models. 

% \chen{Please also tell how you manually feed the (how many) images for generating the adversarial examples}
For each model, according to its functionality, we manually find 10 random images from the Internet as the original input.
For instance, we find 10 images, including 10 different kinds of plants for model 1. 
These images are used by ModelAttacker to generate adversarial examples, which are then used against the selected 10 models.
Note that since finding images for each model takes much human effort, we limit this experiment to 10 models and 10 images for each model, which balances the result validity and human effort.

\subsubsection{Evaluation Metrics}
We evaluate results from two aspects: whether the adversarial examples can be recognized by humans and the attack's success rate.
During the attack, ModelAttacker continuously adjusts $\epsilon$ to determine the most significant attack parameters.
After attacking, two researchers with experience in machine learning made a manual evaluation of the attack results, respectively. 
% They analyzed one by one whether human beings could easily recognize the adversarial examples of successful attacks. 
If the adversarial example cannot be recognized as modified by humans, the attack is considered successful, and we will count the number of examples that successfully misjudged the model among these 10 adversarial examples. 
Otherwise, the attack fails too.
Let $n_i$ represents $n$ examples are misclassified in $i_{th}$ attack, so
\begin{equation}
    P_i = \frac{n_i}{m}
\end{equation}
where $m$ is equal to 10, which is the total number of adversarial examples, and $P_i$ represents the success rate of the $i_{th}$ adversarial attack.

Besides, to evaluate the effectiveness of the approach, we add one control experiment. 
In the comparative experiment, ModelAttacker is unknown about the fine-tuned model's pre-trained model and directly carried out blind attacks trained on randomly selected pre-trained models.
% \chen{I do not think the last sentence is correct. It is just blind attack which is trained on random selected model, right? Please check with Yujin}
Since the adversarial attack is trained on a random pre-trained DL model, we call it \textit{blind attack} for brevity.
In comparison, we call the attack with knowing the pre-trained model as \textit{targeted attack}.

%\subsubsection{Limitation of ModelAttacker} Current approach has two obvious defects. One is that you cannot attack a model that has been encrypted, the other is that the success rates of attacks on some models are very low and requires a lot of time.

\subsection{Results}

% Table generated by Excel2LaTeX from sheet 'attack'
\begin{table*}[htbp]
  \centering
  \caption{Results of targeted and blind attacks}
  \scalebox{1}{
    \begin{tabular}{cc|cc|cc|cc|cc|cc|cc|cc|cc|cc|cc|cc}
    \noalign{\hrule height 1pt}
    \textbf{Attack} & \textbf{Epsilon} & \multicolumn{2}{c|}{\textbf{M1}} & \multicolumn{2}{c|}{\textbf{M2}} & \multicolumn{2}{c|}{\textbf{M3}} & \multicolumn{2}{c|}{\textbf{M4}} & \multicolumn{2}{c|}{\textbf{M5}} & \multicolumn{2}{c|}{\textbf{M6}} & \multicolumn{2}{c|}{\textbf{M7}} & \multicolumn{2}{c|}{\textbf{M8}} & \multicolumn{2}{c|}{\textbf{M9}} & \multicolumn{2}{c|}{\textbf{M10}} & \multicolumn{2}{c}{\textbf{Average}}\\
    \noalign{\hrule height 1pt}
     & & \textbf{T} & \textbf{B} & \textbf{T} & \textbf{B} & \textbf{T} & \textbf{B} & \textbf{T} & \textbf{B} & \textbf{T} & \textbf{B} & \textbf{T} & \textbf{B} & \textbf{T} & \textbf{B} & \textbf{T} & \textbf{B} & \textbf{T} & \textbf{B} & \textbf{T} & \textbf{B} & \textbf{T} & \textbf{B}\\
    \hline
    Boundary Attack & 2.5   & {0.1} & 0.1   & {0} & 0     & {0} & 0     & \textbf{0.1} & 0     & {0} & 0     & {0} & 0     & \textbf{0.1} & 0     & {0} & 0     & {0.2} & 0.2   & {0.3} & 0.3   & \textbf{0.08}  & 0.06 \\
    DDN Attack & 0.5   & {0.1} & 0.1   & {0} & 0     & \textbf{0.1} & 0     & \textbf{0.1} & 0     & {0} & 0     & {0} & 0     & \textbf{0.1} & 0     & {0} & 0     & {0.2} & 0.2   & {0.3} & 0.3   & \textbf{0.09}  & 0.06 \\
    L2 DeepFool Attack & 1.4   & {0.1} & 0.1   & {0} & 0     & \textbf{0.1} & 0     & \textbf{0.2} & 0     & {0} & 0     & {0} & 0     & {0} & 0     & {0} & 0     & {0.2} & 0.2   & {0.3} & 0.3   & \textbf{0.09}  & 0.06 \\
    FGSM  & 0.02  & \textbf{0.4} & 0.2   & {0} & 0     & \textbf{0.1} & 0     & \textbf{0.3} & 0     & \textbf{0.1} & 0     & {0} & 0     & \textbf{0.3} & 0     & {0} & 0     & \textbf{0.7} & 0.2   & \textbf{0.8} & 0.3   & \textbf{0.27}  & 0.07 \\
    Inversion Attack & 10    & \textbf{0.6} & 0.1   & \textbf{0.3} & 0     & \textbf{0.2} & 0     & \textbf{0.4} & 0     & \textbf{0.2} & 0     & \textbf{0.3} & 0     & \textbf{0.3} & 0     & \textbf{0.1} & 0     & \textbf{0.7} & 0.2   & \textbf{0.9} & 0.4   & \textbf{0.4}   & 0.07 \\
    L2 BIM & 1     & {0.1} & \textbf{0.2}   & {0} & 0     & \textbf{0.1} & 0     & \textbf{0.2} & 0     & {0} & 0     & \textbf{0.1} & 0     & \textbf{0.2} & 0.1   & {0} & 0     & \textbf{0.3} & 0.2   & \textbf{0.7} & 0.3   & \textbf{0.17}  & 0.08 \\
    L2 PGD & 12    & {0.2} & 0.2   & {0} & 0     & \textbf{0.1} & 0     & \textbf{0.3} & 0     & \textbf{0.1} & 0     & \textbf{0.1} & 0     & \textbf{0.5} & 0     & \textbf{0.2} & 0     & \textbf{0.7} & 0.2   & \textbf{0.7} & 0.3   & \textbf{0.29}  & 0.07 \\
    Linf BIM & 0.05  & \textbf{0.5} & 0.2   & \textbf{0.1} & 0     & \textbf{0.3} & 0     & \textbf{0.5} & 0     & \textbf{0.1} & 0     & \textbf{0.1} & 0     & \textbf{0.2} & 0     & \textbf{0.5} & 0     & \textbf{0.7} & 0.2   & \textbf{0.9} & 0.3   & \textbf{0.39}  & 0.07 \\
    Linf PGD & 0.05  & \textbf{0.4} & 0.2   & {0} & 0     & \textbf{0.1} & 0     & \textbf{0.2} & 0     & \textbf{0.1} & 0     & \textbf{0.2} & 0     & \textbf{0.3} & 0.1   & \textbf{0.1} & 0     & \textbf{0.7} & 0.2   & \textbf{0.7} & 0.3   & \textbf{0.28}  & 0.08 \\
    Newton Fool Attack & 12    & \textbf{0.4} & 0.1   & {0} & 0     & \textbf{0.1} & 0     & \textbf{0.3} & 0     & \textbf{0.1} & 0     & {0} & 0     & \textbf{0.4} & 0     & \textbf{0.1} & 0     & \textbf{0.3} & 0.2   & \textbf{0.4} & 0.3   & \textbf{0.21}  & 0.06 \\
    Salt and Pepper Noise Attack & 80    & {0.1} & 0.1   & {0} & 0     & \textbf{0.1} & 0     & \textbf{0.3} & 0     & \textbf{0.2} & 0     & {0} & 0     & \textbf{0.7} & 0     & \textbf{0.1} & 0     & \textbf{0.3} & 0.2   & {0.3} & 0.3   & \textbf{0.21}  & 0.06 \\
    \noalign{\hrule height 1pt}
    Average (Models) &       & \textbf{0.27} & 0.15  & \textbf{0.04} & 0     & \textbf{0.12} & 0     & \textbf{0.26} & 0     & \textbf{0.08} & 0     & \textbf{0.07} & 0     & \textbf{0.28} & 0.02  & \textbf{0.1} & 0     & \textbf{0.45} & 0.2   & \textbf{0.57} & 0.31  & \textbf{0.23}  & 0.07 \\
    \noalign{\hrule height 1pt}
    \end{tabular}%
    }
  \label{tab:results}%
\end{table*}%

\begin{comment}
Table~\ref{tab:results} shows the results of selected 11 adversarial attacks, where \emph{Ave (M)} means the average success rates of ModelAttacker, \emph{Ave (U)} means the average success rates of the traditional approach, and \emph{Ave (A)} represents the average success rates of ModelAttacker and the traditional approach. 
\chen{Do not need the abbreviation if you just use it once,.}
The \emph{MobilenetV1} and \emph{MobilenetV2} represent that they has been confirmed that their pre-training models are MobilenetV1 or V2 and attacked by ModelAttacker. 
In contrast, \emph{Unknown pre-trained models} represent not knowing the pre-training model and is attacked by traditional methods.
\end{comment}

\begin{comment}
To answer the question 'Q3: How robust are fine-tuning TFLite DL models against adversarial attacks ?', we will evaluate the robust by the success rate of each attack, and compare the results of ModelAttacker with traditional adversarial attacks.
The higher the attack success rate, the easier the models will be attacked.
Then we will compare the robust of three fine-tuning methods feature extraction, partial freezing and big weight change.
Finally, we will compare the effect of every type of adversarial attacks.
\chen{Just analyze results, not giving conclusions. Please use the data to in the text like is the similarity of model will influence the success attack rate?}
\end{comment}

Table~\ref{tab:results} shows the comparative results of targeted adversarial attack and blind adversarial attack.
In every model's column, the left side \emph{T} is the result of targeted attacks, and the right side \emph{B} is the result of blind attacks. The better results are blackened. The last two columns are the average success rate of targeted and blind attacks.
Compared with the blind adversarial attacks trained on random models, the targeted attacks trained on pre-trained models have much better success rates.
On average, blind adversarial attacks can achieve a 0.07 success rate, while the targeted attacks' success rate is  0.23, 229\% higher than the blind attack.
The blind adversarial attacks do not work at all in 6 models, including model 2, 3, 4, 5, 6, and 8, with a success rate of 0 given all 11 attacks.
Instead, our targeted attacks work on all models with the average success rate ranging from 0.08 to 0.57, and some attack model can achieve even as high as 0.9 success rate like Linf BIM attack on model 10.
Within 110 experiments (10 models * 11 attacks), the targeted attack gets a higher or equal success rate in 99\% of them. 
These results demonstrate the necessity of locating a pre-trained model before carrying out the adversarial attack.
\begin{comment}
First, we can see from Table~\ref{tab:results} that when using ModelAttacker, model 10 has the highest average success rate, reaching 0.57. 
For the traditional approach, model 10 has the highest average success rate too. 
However it is only 0.31, which is 45.67\% lower than when ModelAttacker is used.
For the ModelAttacker, the success rate of Model 2 is the lowest, only 0.04, but it is still higher than the traditional approach, because model 2, 3, 4, 5 and 6 are all 0 in the traditional approach.
Besides, ModelAttacker's average success rates of every model are all higher than that in the traditional approach. 
For example, the average success rate of model 7 has increased from 0.02 to 0.28, an increase of 14 times. 
The average success rate of model 4 has increased from 0 to 0.26, making the model considered safe in traditional methods unsafe.
In total, 6 of 10 models have 0 success rates in the traditional approach, which means these 6 models can defend against adversarial attacks. 
In contrast, ModelAttacker makes all ten models be attacked successfully.
Compared with the traditional approach, ModelAttacker attacks the fine-tuning model's pre-trained model in advance and obtains the most significant attack parameters and adversarial examples. 
Therefore, some fine-tuning model structure, transferred from the pre-trained model, is fully exposed to ModelAttacker.
ModelAttacker already knows how to attack this exposed part of the model, so the attacks' success rate can be significantly improved.
\end{comment}

\begin{figure}
    \centering
    \includegraphics[width=0.9\linewidth]{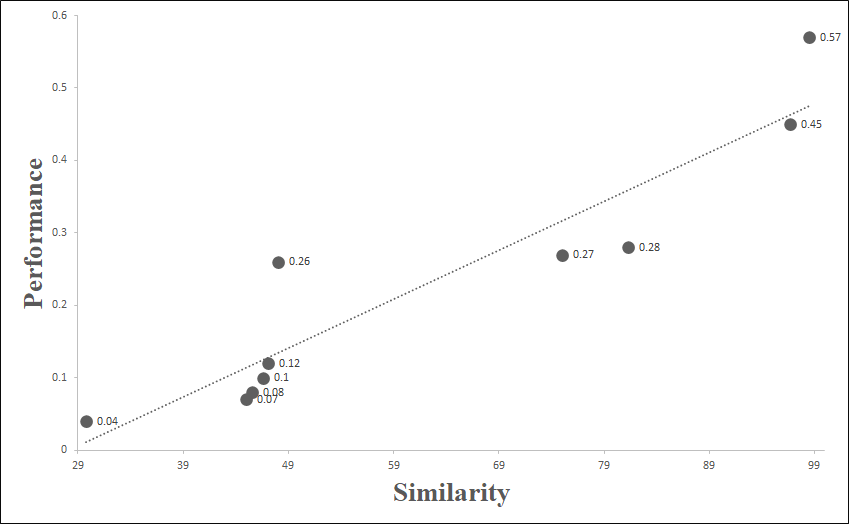}
    \caption{Relationship of similarity and attack performance (the dot line is the fitted line)}
    \label{fig:scatter plot}
\end{figure}

Some deep learning models are more vulnerable to adversarial attacks than others.
As seen in Table~\ref{tab:results}, models 9 and 10 are easier to be attacked than other models, and almost all targeted attack methods work well on them.
The reason may be that they are heavily relying on pre-trained models as feature extractors.
Table~\ref{tab:model_detail} shows that mode 9 and 10 are more than 90\% similar to the provided MobileNetV2 by Google in TensorHub, and only add a classification layer at the end of the MobileNetV2.
Since our targeted attack is trained on the pre-trained model, hence with good performance (\emph{Average(Models)}) on them.
To further explore the relationship between the model similarity and attack performance, we draw a scatter plot on Figure~\ref{fig:scatter plot} to show their relationship and also carry out the Pearson correlation coefficient (PCC)~\cite{benesty2009pearson} to check if the correlation exists between them.
The results show that their PCC is 0.93, which means they are positive linear correlation in current data.
This similarity ranges from 50 to 80, and the performance grows very slowly. In contrast, performance grows rapidly in the 40 to 50 and 80 to 100 ranges.

Some attack models (e.g., Inversion Attack, Linf Bim, L2 PGD) are more effective than others (e.g., Boundary Attack, DDN Attack).
In particular, our experiments in Table~\ref{tab:results} show that the \emph{Inversion Attack} is the most threading adversarial attack, with an average success rate of 0.4.
Different from other models, it works in all 10 models in our study.
%\chen{Please explain the potential reasons why it is very effective, and I cannot understand what you write below}
Inversion attack takes the confidence of a prediction result in training data as the target. It adopts gradient descent to revise the adversarial examples according to the prediction result repeatedly and continuously improve the adversarial examples' prediction confidence. Targeted attacks more clearly point out the direction of optimizing the adversarial examples for the attack, so it can improve the attack's success rate more effectively. 

% Compared with other attacks, such as \emph{inversion attack} and \emph{BIM attack}, \emph{boundary attack} do not make full use of the information obtained by ModelAttacker, but only use the target model's prediction of the sample, so the effect of this attack method is not much improved. And compared with other attacks, it requires more attack time.

%Third, it can also be seen from Table~\ref{tab:results} that the \emph{Inversion Attack} is the most threading adversarial attack, with an average success rate of 0.235. The \emph{Boundary Attack} is the weakest attack, with an average success rate of 0.07. 

\begin{center}
\begin{tcolorbox}[colback=black!5!white,colframe=black!75!black]
\textbf{Summary:} ModelAttacker can significantly improve the adversarial attacks' success rate by identifying the pre-trained model compared with the blind attack.
The more similar is the target model to the pre-trained model, the more vulnerable it is to our targeted adversarial attacks.
The attack methods like inversion attack are more effective than others in targeted attacks.
%Besides, the more similar the fine-tuning model is to its pre-training model, the more information it exposes through the pre-training model, and the higher the success rate of adversarial attacks.
\end{tcolorbox}
\end{center}

\section{Discussion}
\label{section:discussion}
% In this section, we will discuss some observations and insights of defending against adversarial attacks and protecting DL model intellectual property (IPs) based on this paper's results.

\subsection{Defence of Attack}
% \chen{Please give the defence suggestion according to the findings of RQ3.}
% Attackers can easily attack the DL apps with similar DL models via adversarial attack\cite{qiu2019review,yuan2019adversarial,zhang2019adversarial}. 
% Large-scale DL model similarity should arouse people's awareness of the security of DL model and the protection of intellectual property of DL model. 

From our investigation in Section~\ref{section:Q2}, most models currently used in Android apps are fine-tuned from several popular pre-trained models, such as MobileNet and DeepLabV3.
Some fine-tuned models' names are even quite similar to the pre-trained models, for instance, \emph{mobilenet.letgo.v1\_1.0\_224\_quant.v7.tflite}, which can be easily assumed that they are fine-tuned from MobileNet.
However, according to Table\ref{tab:results}, the targeted attacks are significantly threatening than the blind attacks as they are generated by attacking the specific pre-trained models. 
Hence, some measures for preventing disclosure of model information are necessary, such as encrypt the pre-trained models, confuse the predicted labels.
However, according to the Section\ref{section:Q3}, the higher the similarity between the fine-tuned model and the pre-trained model, the easier it is to be attacked.
Thus, we should change both the structure and parameters of the pre-trained model as much as possible, especially when using common pre-trained models, such as MobileNet and DeepLabV3.

% Furthermore, we find that except for a few Android apps that have implemented protection measures, such as model file encryption, the models in most Android apps are directly exposed to attackers at present. 
% A model protection mechanism is compulsory. 
% So, how to design an effective model protection mechanism will be an auspicious research direction.

According to the prediction results, some attacks, such as the inversion attack, continuously revises adversarial examples to improve the prediction confidence. We should take measures, such as fuzzy confidential information~\cite{fussy_confidence} and differential privacy technology~\cite{differential_privacy}. Let the attacker be impossible to detect each training data change's impact on the model output, reducing these attacks' threat.

\subsection{Deep Learning Model IP}
% Academia and industry have spent a lot of resources to develop deep-learning-based apps or embed deep learning modules into mobile apps. 
% However, 
Building a product-level deep learning model is not straightforward, as it always requires lots of computing resources and training data. 
For instance, XL-net\cite{XL-net}, which is proposed by researchers from Google AI Brain, utilizes over 120GB of training data and 512 TPU v3 chips to train the model. 
Furthermore, designing model architectures and selecting hyper-parameters are time-consuming. 
Some mobile developers may illegally infringe intellectual property by reusing others' deep learning model file or retraining a new deep learning model on top of them.
Hence, model owners also possess the IP of their trained models. Unfortunately, out of 120 popular deep learning apps, only 47 (39.2\%) of model owners take measures to protect their IP\cite{xu2019first}. 
In addition, at present only a few deep learning R\&D frameworks,  such as ncnn\cite{ncnn} and Mace\cite{mace}, provide the function of protecting the IP of deep learning models. 
We should take the necessary measures to protect the IP of the Android phone's deep learning model, such as watermarking, file encryption, etc.
Thus, protecting the IPs of deep learning models on smartphones is worth exploring, and call on researchers and developers to pay more attention to this field.

% and has become one of the most vital research trends as more and more deep learning models are deployed on smartphones.

\section{Related Works}
\label{related work}

With the rapid development of computing architectures, hardware like GPU has been released at affordable prices, facilitating the deployment of deep learning models on mobile devices\cite{ota2017deep}. Our work empirically evaluates the security of deep learning models on Android apps. It interacts with three lines of research: mobile app security, adversarial deep learning, and proprietary model protection.

\subsection{Mobile App Security}

Prior work on AI app security mainly focuses on model protection. Xu et al.\cite{xu2019first} indicate that most deep learning models are exposed without protection in deep learning apps, so can easily be extracted and exploited by attackers. Sun et al.\cite{sun2020mind} demonstrate the feasibility of private model extraction on AI apps, subsequently, they further analyse the model parameters and show the estimated financial loss caused by the leaked models. Wang et al.\cite{wang2018deep} comprehensively investigate the current challenges encountered when pushing deep learning towards mobile apps, and find that model protection is crucial, as leaking critical models can result in both security and privacy breaches. These studies explore the security of deep learning models either from the user's or the defender's point of view, in lack of real attacks. Hence, motivated by enormous efforts to secure AI apps, our work investigates the security of deep learning models from the attacker's perspective, providing a valuable method with which to test AI app security.

\subsection{Adversarial Attack and Defence to Deep Learning Models}

Since deep learning has achieved remarkable progress in the field of computer vision, main developments in adversarial attack and defence are associated with image classification\cite{szegedy2013intriguing,goodfellow2014explaining,carlini2017towards,deepbackdoor}. Researchers have
proposed numerous novel adversarial attacks based on deep neural networks\cite{goodfellow2014explaining,dong2018boosting,carlini2017towards}, which can be mainly summarized into two categories including white-box and black-box attacks\cite{he2019towards}. For example, Szegedy et al.\cite{szegedy2013intriguing} propose an optimisation function to construct adversarial examples, solving it with L-BFGS. To prevent the aforementioned
attacks, seven defence methods have been proposed, including adversarial training, transformation, distillation and gradient regularisation\cite{he2019towards}. Among these countermeasures, adversarial training draws the most attention and it is one of the most effective defence methods\cite{zhang2019defense}. Despite numerous studies that have proposed various adversarial attack methods and corresponding defence strategies for deep learning models, work on the security of deep learning models on mobile apps is scant. Our work proves that most on-device deep learning models lack protection against adversarial attacks, especially in the field of computer vision.

\begin{comment}
\subsection{Deep Learning Model Intellectual Property}

To protect the DL model as an intellectual property, Several watermarking mechanisms\cite{adi2018turning,nagai2018digital,rouhanideepsigns,zhang2018protecting} have been proposed to detect illegitimate model uses. Recently, Chakraborty et al.\cite{chakraborty2020hardware} proposed a hardware root-of-trust named HPNN to safeguard the intellectual property of DL models that potential attackers have access to. However, These approaches cannot completely prevent DL models from extraction and attack. Our work proves that model plagiarism is a realistic problem in DL apps especially for Android platform.
\end{comment}
\section{Conclusion}
This paper proposes a practical approach to hacking deep learning models with adversarial attacks by identifying highly similar pre-trained models from TensorFlow Hub.
% We build ModelAttacker to perform adversarial attacks on 10 representative fine-tuned TFLite models used in Android apps for experiments. 
Experimental results show that attacks with knowing fine-tuned models' pre-trained models are more than 3 times better than blind attacks in terms of the attack success rate.
The results also demonstrate that the higher the similarity between the fine-tuned model and its pre-training model, the worse the fine-tuned model's robustness. 
In addition, we carry out an empirical study into the use of deep learning models in real-world Android apps, identifying 267 pre-trained TFLite models among 62,822 Android apps. 
%87.64\% of 267 extracted models are structurally similar, and 79.78\% of them have the same parameters as at least one model.
Excluding 182 identical models, we find that 61.18\% of the rest are more than 80\% similar to pre-trained models from TensorFlow Hub in terms of the structural similarity. 
% 50.59\%of them have more than 43.93\% same parameters from which feature extraction and fine-tuning technique is widely used to those pre-trained models.

%Finally, we discuss some suggestions about defending against adversarial attacks and call to protect DL models' IP.  
In the future, we will work in two directions in the fields of both attack and defence.
We will develop specific adversarial attacks that target pre-trained models, thereby making the attack more effective.
Additionally, we will try to implement the defence mechanism in practice, in order to establish how to best protect on-app deep learning models.

% \chen{Please update this part (especially the data) by referring to the new abstract and introduction.}

\label{conclusion}

\bibliography{reference}
\bibliographystyle{abbrv}

\end{document}